\DocumentMetadata{
  lang        = en,
  pdfversion  = 2.0,
  pdfstandard = ua-2,
  pdfstandard = a-4,
  testphase   = latest
}

\documentclass[numsec,webpdf,modern,large]{oup-authoring-template}

\graphicspath{{Fig/}}

\usepackage{colortbl}
\usepackage{booktabs}
\usepackage{bm}

\theoremstyle{thmstyleone}%
\theoremstyle{thmstyletwo}%
\theoremstyle{thmstylethree}%

\begin{document}

\journaltitle{Journal Title Here}
\DOI{DOI HERE}
\copyrightyear{2022}
\pubyear{2019}
\access{Advance Access Publication Date: Day Month Year}
\appnotes{Paper}

\firstpage{1}


\title[\textsc{FlowDock}: Geometric Flow Matching for Generative Protein-Ligand Docking and Affinity Prediction]{\textsc{FlowDock}: Geometric Flow Matching for Generative Protein-Ligand Docking and Affinity Prediction}

\author[1,$\ast$]{Alex Morehead\ORCID{0000-0002-0586-6191}}
\author[1]{Jianlin Cheng\ORCID{0000-0003-0305-2853}}

\authormark{Morehead et al.}

\address[1]{\orgdiv{Department of Electrical Engineering \& Computer Science, NextGen Precision Health}, \orgname{University of Missouri-Columbia}, \orgaddress{\street{W1024 Lafferre Hall}, \postcode{65211}, \state{Missouri}, \country{USA}}}

\corresp[$\ast$]{Corresponding author. \href{email:acmwhb@missouri.edu}{acmwhb@missouri.edu}}

\received{Date}{0}{Year}
\revised{Date}{0}{Year}
\accepted{Date}{0}{Year}



\abstract{
\textbf{Motivation}\\
Powerful generative AI models of protein-ligand structure have recently been proposed, but few of these methods support both flexible protein-ligand docking and affinity estimation. Of those that do, none can directly model multiple binding ligands concurrently or have been rigorously benchmarked on pharmacologically relevant drug targets, hindering their widespread adoption in drug discovery efforts.\\
\textbf{Results}\\
In this work, we propose \textsc{FlowDock}, the first deep geometric generative model based on conditional flow matching that learns to directly map unbound (apo) structures to their bound (holo) counterparts for an arbitrary number of binding ligands. Furthermore, \textsc{FlowDock} provides predicted structural confidence scores and binding affinity values with each of its generated protein-ligand complex structures, enabling fast virtual screening of new (multi-ligand) drug targets. For the well-known PoseBusters Benchmark dataset, \textsc{FlowDock} outperforms single-sequence AlphaFold 3 with a 51\% blind docking success rate using unbound (apo) protein input structures and without any information derived from multiple sequence alignments, and for the challenging new DockGen-E dataset, \textsc{FlowDock} outperforms single-sequence AlphaFold 3 and matches single-sequence Chai-1 for binding pocket generalization. Additionally, in the ligand category of the 16th community-wide Critical Assessment of Techniques for Structure Prediction (CASP16), \textsc{FlowDock} ranked among the top-5 methods for pharmacological binding affinity estimation across 140 protein-ligand complexes, demonstrating the efficacy of its learned representations in virtual screening. \\
\textbf{Availability}\\
Source code, data, and pre-trained models are available at \url{https://github.com/BioinfoMachineLearning/FlowDock}.
}

\maketitle

\section{Introduction}
Interactions between proteins and small molecules (ligands) drive many of life's fundamental processes and, as such, are of great interest to biochemists, biologists, and drug discoverers. Historically, elucidating the structure, and therefore the function, of such interactions has required that considerable intellectual and financial resources be dedicated to determining the interactions of a single biomolecular complex. For example, techniques such as X-ray diffraction and cryo-electron microscopy have traditionally been effective in biomolecular structure determination, however, resolving even a single biomolecule's crystal structure can be extremely time and resource-intensive. Recently, new machine learning (ML) methods such as AlphaFold 3 (AF3) \citep{abramson2024accurate} have been proposed for directly predicting the structure of an arbitrary biomolecule from its primary sequence, offering the potential to expand our understanding of life's molecules and their implications in disease, energy research, and beyond.

Although powerful models of general biomolecular structure are compelling, they currently do not provide one with an estimate of the binding affinity of a predicted protein-ligand complex, which may indicate whether a pair of molecules truly bind to each other \textit{in vivo}. It is desirable to predict both the structure of a protein-ligand complex and the binding affinity between them via one single ML system \citep{dhakal2022artificial}.  Moreover, recent generative models of biomolecular structure are primarily based on noise schedules following Gaussian diffusion model methodology which, albeit a powerful modeling framework, lacks interpretability in the context of biological studies of molecular interactions. In this work, we aim to address these concerns with a new \textit{state-of-the-art} hybrid (structure \& affinity prediction) generative model called \textsc{FlowDock} for flow matching-based protein-ligand structure prediction and binding affinity estimation, which allows one to interpretably inspect the model's structure prediction trajectories to interrogate its common molecular interactions and to screen drug candidates quickly using its predicted binding affinities.

\section{Related work}
\textbf{Molecular docking with deep learning.} Over the last few years, deep learning (DL) algorithms (in particular geometric variants) have emerged as a popular methodology for performing end-to-end differentiable molecular docking. Models such as EquiBind \citep{stark2022equibind} and TankBind \citep{lu2022tankbind} initiated a wave of interest in researching graph-based approaches to modeling protein-ligand interactions, leading to many follow-up works. Important to note is that most of such DL-based docking models were designed to supplement conventional modeling methods for protein-ligand docking such as AutoDock Vina \citep{eberhardt2021autodock} which are traditionally slow and computationally expensive to run for many protein-ligand complexes yet can achieve high accuracy with crystal input structures and ground-truth binding pocket annotations. \\
\textbf{Generative biomolecular modeling.} The potential of generative modeling in capturing intricate molecular details in structural biology such as protein-ligand interactions during molecular docking \citep{corso2022diffdock} has recently become a research focus of ambitious biomolecular modeling efforts such as AF3 \citep{abramson2024accurate}, with several open-source spin-offs of this algorithm emerging \citep{chai2024chai, wohlwend2024boltz}.\\
\textbf{Flow matching.} In the machine learning community, generative modeling with flow matching \citep{chen2024flow, tong2024improving} has recently become an appealing generalization of diffusion generative models \citep{ho2020denoising, karras2022elucidating}, enabling one to transport samples between arbitrary distributions for compelling applications in computer vision \citep{esser2024scaling}, computational biology \citep{klein2024equivariant}, and beyond. As a closely related concurrent work (as our method was developed for the CASP16 competition starting in May 2024 \citep{casp16_abstracts}), \cite{corso24flexible} recently introduced and evaluated an unbalanced flow matching procedure for pocket-based flexible docking. However, the authors' proposed approach mixes diffusion and flow matching noise schedules with geometric product spaces in an unintuitive manner, and neither source code nor data for this work are publicly available for benchmarking comparisons. In Section \ref{section:riemannian_manifolds_and_conditional_flow_matching}, we describe flow matching in detail.\\
\textbf{Contributions.} In light of such prior works, our contributions in this manuscript are as follows:

\begin{itemize}
    \item We introduce the \textit{first} simple yet state-of-the-art \textit{hybrid} generative flow matching model capable of quickly and accurately predicting protein-ligand complex structures \textit{and} their binding affinities, with source code and model weights freely available.
    \item We rigorously validate our proposed methodology using standardized benchmarking data for protein-ligand complexes, with our method ranking as a more accurate and generalizable structure predictor than (single-sequence) AF3.
    \item Our method ranked as a top-5 binding affinity predictor for the 140 pharmaceutically relevant drug targets available in the 2024 community-wide CASP16 ligand prediction competition.
    \item We release one of the largest ML-ready datasets of apo-to-holo protein structure mappings based on high-accuracy predicted protein structures, which enables training new models on comprehensive biological data for distributional biomolecular structure modeling.
\end{itemize}

\section{Methods and materials}
The goal of this work is to jointly predict protein-ligand complex structures and their binding affinities with minimal computational overhead to facilitate drug discovery. In Sections \ref{section:overview} and \ref{section:notation}, we briefly outline how \textsc{FlowDock} achieves this and how its key notation is defined. We then describe \textsc{FlowDock}'s training and sampling procedures in Sections \ref{section:riemannian_manifolds_and_conditional_flow_matching}-\ref{section:sampling}.

\subsection{\textsc{Overview}}
\label{section:overview}

\begin{figure*}[t]
    \centering
    \includegraphics[width=\textwidth, alt={An overview of biomolecular distribution modeling with \textsc{FlowDock}.}]{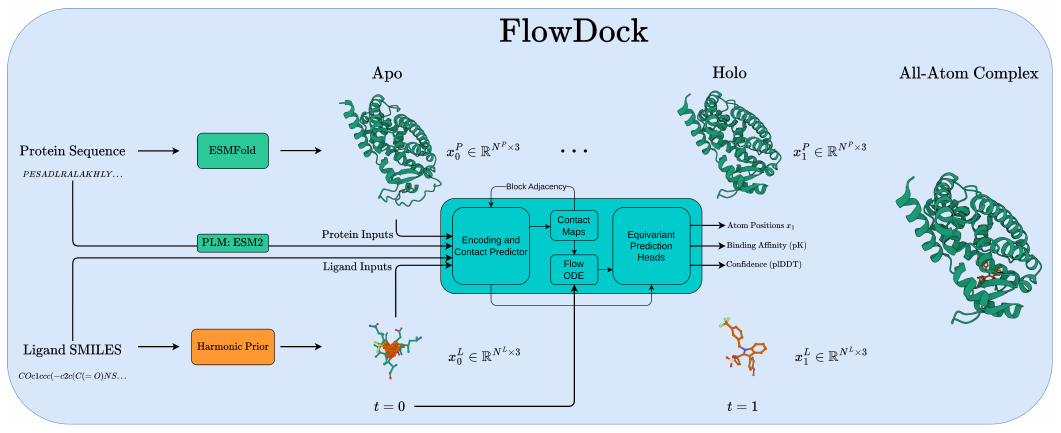}
    \caption{An overview of biomolecular distribution modeling with \textsc{FlowDock}.}
    \label{figure:flowdock}
\end{figure*}

Figure \ref{figure:flowdock} illustrates how \textsc{FlowDock} uses geometric flow matching to predict flexible protein-ligand structures and binding affinities. At a high level, \textsc{FlowDock} accepts both (multi-chain) protein sequences and (multi-fragment) ligand SMILES strings as its primary inputs, which it uses to predict an unbound (apo) state of the protein sequences using ESMFold \citep{lin2023evolutionary} and to sample from a harmonic ligand prior distribution \citep{jingalphafold} to initialize the ligand structures using biophysical constraints based on their specified bond graphs. Notably, users can also specify the initial protein structure using one produced by another bespoke method (e.g., AF3 which we use in certain experiments). With these initial structures representing the complex's state at time $t=0$, \textsc{FlowDock} employs conditional flow matching to produce fast structure generation trajectories. After running a small number of integration timesteps (e.g., 40 in our experiments), the complex's state arrives at time $t=1$, i.e., the model's estimate of the bound (holo) protein-ligand heavy-atom structure. At this point, \textsc{FlowDock} runs confidence and binding affinity heads to predict structural confidence scores (i.e., plDDT) and binding affinities of the predicted complex structure, to rank-order the model's generated samples.

\subsection{Notation}
\label{section:notation}
Let $\bm{x}_{0}$ denote the unbound (apo) state of a protein-ligand complex structure, representing the heavy atoms of the protein and ligand structures as $\bm{x}_{0}^{P} \in \mathbb{R}^{N^{P} \times 3}$ and $\bm{x}_{0}^{L} \in \mathbb{R}^{N^{L} \times 3}$, respectively, where $N^{P}$ and $N^{L}$ are the numbers of protein and ligand heavy atoms. Similarly, we denote the corresponding bound (holo) state of the complex as $\bm{x}_{1}$. Further, let $\bm{s}^{P} \in \{1,\dots,20\}^{S^{P}}$ denote the type of each amino acid residue in the protein structure, where $S^{P}$ represents the protein's sequence length. To generate bound (holo) structures, we define a flow model $v_{\theta}$ that integrates the ordinary differential equation (ODE) it defines from time $t=0$ to $t=1$.

\subsection{Riemannian manifolds and conditional flow matching}
\label{section:riemannian_manifolds_and_conditional_flow_matching}
In manifold theory, an $n$-dimensional manifold $\mathcal{M}$ represents a topological space equivalent to $\mathbb{R}^{n}$. In the context of Riemannian manifold theory, each point $\bm{x} \in \mathcal{M}$ on a Riemannian manifold is associated with a tangent space $\mathcal{T}_{\bm{x}}$. Conveniently, a Riemannian manifold is equipped with a metric $g_{\bm{x}}: \mathcal{T}_{\bm{x}}\mathcal{M} \times \mathcal{T}_{\bm{x}}\mathcal{M} \rightarrow \mathbb{R}$ that permits the definition of geometric quantities on the manifold such as distances and geodesics (i.e., shortest paths between two points on the manifold). Subsequently, Riemannian manifolds allow one to define on them probability densities $\int_{\mathcal{M}}\rho(\bm{x})d\bm{x} = 1$ where $\rho: \mathcal{M} \rightarrow \mathbb{R}_{+}$ are continuous, non-negative functions. Such probability densities give rise to interpolative probability paths $\rho_{t}: [0, 1] \rightarrow \mathbb{P}(\mathcal{M})$ between probability distributions $\rho_{0}, \rho_{1} \in \mathbb{P}(\mathcal{M})$, where $\mathbb{P}(\mathcal{M})$ is defined as the space of probability distributions on $\mathcal{M}$ and the interpolation in probability space between distributions is indexed by the continuous parameter $t$.

Here, we refer to $\psi_{t}: \mathcal{M} \rightarrow \mathcal{M}$ as a \textit{flow} on $\mathcal{M}$. Such a flow serves as a solution to the ODE: $\frac{d}{dt}\psi_{t}(\bm{x}) = u_{t}(\psi_{t}(\bm{x}))$ \citep{mathieu2020riemannian} which allows one to \textit{push forward} the probability trajectory $\rho_{0} \rightarrow \rho_{1}$ to $\rho_{t}$ using $\psi_{t}$ as $\rho_{t} = [\psi_{t}]_{\#}(\rho_{0})$, with $\psi_{0}(\bm{x}) = \bm{x}$ for $u: [0, 1] \times \mathcal{M} \rightarrow \mathcal{M}$ (i.e., a smooth time-dependent vector field \citep{bosese}). This insight allows one to perform \textit{flow matching} (FM) \citep{chen2024flow} between $\rho_{0}$ and $\rho_{1}$ by learning a continuous normalizing flow \citep{papamakarios2021normalizing} to approximate the vector field $u_{t}$ with the parametric $v_{\theta}$. With $\rho_{0} = \rho_{prior}$ and $\rho_{1} = \rho_{data}$, we have that $\rho_{t}$ advantageously permits \textit{simulation-free} training. Although it is not possible to derive a closed form for $u_{t}$ (which generates $\rho_{t}$) with the traditional flow matching (FM) training objective, a \textit{conditional} flow matching (CFM) training objective remains tractable by marginalizing conditional vector fields as $u_{t}(\bm{x}) := \int_{\mathcal{M}} u_{t}(\bm{x} | \bm{z}) \frac{\rho_{t}(\bm{x}_{t} | \bm{z}) q(\bm{z})}{\rho_{t}(\bm{x})} d\bm{z}$, where $q(\bm{z})$ represents one's chosen coupling distribution (by default the independent coupling $q(\bm{z}) = q(\bm{x}_{0})q(\bm{x}_{1})$) between $\bm{x}_{0}$ and $\bm{x}_{1}$ via the conditioning variable $\bm{z}$. For Riemannian CFM (RCFM) \citep{chen2024flow}, the corresponding training objective, with $t \sim \mathcal{U}(0, 1)$, is:
\begin{equation}
    \label{equation:rcfm_loss}
    \mathcal{L}_{RCFM}(\theta) = \mathbb{E}_{t, q(\bm{z}), \rho_{t}(\bm{x}_{t} | \bm{z})} \lVert v_{\theta}(\bm{x}_{t}, t) - u_{t}(\bm{x}_{t} | \bm{z}) \rVert_{g}^{2},
\end{equation}
where \citet{tong2024improving} have fortuitously shown that the gradients of FM and CFM are identical. As such, to transport samples of the prior distribution $\rho_{0}$ to the target (data) distribution $\rho_{1}$, one can sample from $\rho_{0}$ and use $v_{\theta}$ to run the corresponding ODE forward in time. In the remainder of this work, we will focus specifically on the 3-manifold $\mathbb{R}^{3}$.

\subsection{Prior distributions}
\label{section:prior_distributions}
With flow matching defined, in this section, we describe how we use a bespoke mixture of prior distributions ($\rho_{0}^{P}$ and $\rho_{0}^{L}$) to sample initial (unbound) protein and ligand structures for binding (holo) structure generation targeting our data distribution of crystal protein-ligand complex structures $\rho_{1}$. In Section \ref{section:posebench_protein_ligand_docking}, we ablate this mixture to understand its empirical strengths.

\textbf{ESMFold protein prior.} To our best knowledge, \textsc{FlowDock} is among the \textit{first} methods-concurrently with \cite{corso24flexible}-to explore using structure prediction models with flow matching to represent the unbound state of an arbitrary protein sequence. In contrast to \cite{corso24flexible}, we formally define a \textit{distribution} of unbound (apo) protein structures using the single-sequence ESMFold model as $\rho_{0}^{P}(\bm{x}_{0}^{P}) \propto \mathrm{ESMFold}(\bm{s}^{P}) + \epsilon, \hspace{0.25cm} \epsilon \sim \mathcal{N}(0, \sigma)$, which encourages our model to learn more than a strict mapping between protein apo and holo point masses. Based on previous works developing protein generative models \citep{dauparas2022robust}, during training we apply $\epsilon \sim \mathcal{N}(0, \sigma=1e^{-4})$ to both $\bm{x}_{0}^{P}$ and $\bm{x}_{1}^{P}$ to discourage our model from overfitting to computational or experimental noise in its training data. It is important to note that this additive noise for protein structures is not a general substitute for generating a full conformational ensemble of each protein, but to avoid the excessively high computational resource requirements of running protein dynamics methods such as AlphaFlow \citep{jingalphafold} for each protein, we empirically find noised ESMFold structures to be a suitable surrogate.

\textbf{Harmonic ligand prior.} Inspired by the FlowSite model for multi-ligand binding site design \citep{starkharmonic}, \textsc{FlowDock} samples initial ligand conformations using a harmonic prior distribution constrained by the bond graph defined by one's specified ligand SMILES strings. This prior can be sampled as a modified Gaussian distribution via $\rho_{0}^{L}(\bm{x}_{0}^{L}) \propto exp(-\frac{1}{2}\bm{x}_{0}^{L^{T}}\bm{L}\bm{x}_{0}^{L})$ where $\bm{L}$ denotes a ligand bond graph's Laplacian matrix defined as $\bm{L} = \bm{D} - \bm{A}$, with $\bm{A}$ being the graph's adjacency matrix and $\bm{D}$ being its degree matrix. Similarly to our ESMFold protein prior, we subsequently apply $\epsilon \sim \mathcal{N}(0, \sigma=1e^{-4})$ to $\bm{x}_{1}^{L}$ during training.

\subsection{Training}
\label{section:training}
We describe \textsc{FlowDock's} structure parametrization, optimization procedure, and the curation and composition of its new training dataset in the following sections. Further, we provide training and inference pseudocode in Appendix \ref{appendix:pseudocode} of our Supplementary Materials.

\textbf{Parametrizing protein-ligand complexes with geometric flows.} Based on our experimental observations of the difficulty of scaling up intrinsic generative models \citep{corso2023modeling} that operate on geometric product spaces, \textsc{FlowDock} instead parametrizes 3D protein-ligand complex structures as attributed geometric graphs \citep{joshi2023expressive} representing the heavy atoms of each complex's protein and ligand structures. The main benefit of a heavy atom parametrization is that it can considerably simplify the optimization of a flow model $v_{\theta}$ by allowing one to define its primary loss function as simply as a CondOT path \citep{pooladian2023multisample, jingalphafold}:
\begin{equation}
    \label{equation:r3_loss}
    \mathcal{L}_{\mathbb{R}^{3}}(\theta) = \mathbb{E}_{t, q(\bm{z}), \rho_{t}(\bm{x}_{t} | \bm{z})} \lVert v_{\theta}(\bm{x}_{t}, t) - \bm{x}_{1}) \rVert^{2},
\end{equation}
with the conditional probability path $\rho_{t}$ chosen as
\begin{equation}
    \label{equation:conditional_probability_path}
    \rho_{t}(\bm{x} | \bm{z}) = \rho_{t}(\bm{x} | \bm{x}_{0}, \bm{x}_{1}) = (1 - t) \cdot \bm{x}_{0} + t \cdot \bm{x}_{1}, \hspace{0.25cm} \bm{x}_{0} \sim \rho_{0}(\bm{x}_{0})
\end{equation}
The challenge introduced by this atomic parametrization is that it necessitates the development of an efficient neural architecture that can scalably process all-atom input structures without the exhaustive computational overhead of generative models such as AF3. Fortunately, one such architecture satisfies this requirement, namely, one recently introduced by \cite{qiao2024state} with the NeuralPLexer model which encodes protein language model (PLM) sequence embeddings and ligand SMILES strings to iteratively decode block diagonal contact maps to condition a flow ODE for equivariant coordinates and auxiliary predictions. As such, inspired by how the AlphaFlow model was fine-tuned from the base AlphaFold 2 (AF2) architecture using flow matching, to train \textsc{FlowDock} we explored fine-tuning the NeuralPLexer architecture to represent our vector field estimate $v_{\theta}$ as illustrated in Figure \ref{figure:flowdock}. Uniquely, we empirically found this idea to work best by fine-tuning the architecture's score head, which was originally trained with a denoising score matching objective for \textit{diffusion}-based structure sampling, instead using Eqs. \ref{equation:r3_loss} and \ref{equation:conditional_probability_path}. Moreover, we fine-tune all of NeuralPLexer's remaining intermediate weights and prediction heads including a dedicated confidence head redesigned to predict binding affinities, with the exception of its original confidence head which remains frozen at all points during training.

\textbf{PDBBind-E Data Curation.} To train \textsc{FlowDock} with resolved protein-ligand structures and binding affinities, we prepared PDBBind-E, an enhanced version of the PDBBind 2020-based training dataset proposed by \cite{corsodeep} for training recent DL docking methods such as DiffDock-L. To curate PDBBind-E, we collected 17,743 crystal complex structures contained in the PDBBind 2020 dataset and 47,183 structures of the Binding MOAD \citep{hu2005binding} dataset splits introduced by \cite{corsodeep} (n.b., which maintain the validity of our benchmarking results in Section \ref{section:results} according to time and ligand-based similarity cutoffs) and predicted the structure of these (multi-chain) protein sequences in each dataset split using ESMFold. To optimally align each predicted protein structure with its corresponding crystal structure, we performed a weighted structural alignment optimizing for the distances of the predicted protein residues' C$\alpha$ atoms to the crystal heavy atom positions of the complex's binding ligand, similar to \citep{corsodeep}. After dropping complexes for which the crystal structure contained protein sequence gaps caused by unresolved residues, the total number of PDBBind and Binding MOAD predicted complex structures remaining was 17,743 and 46,567, respectively.

\textbf{Generalized unbalanced flow matching.} We empirically observed the challenges of naively training flexible docking models like \textsc{FlowDock} without any adjustments to the sampling of their training data. Accordingly, we concurrently developed a generalized version of \textit{unbalanced} flow matching \citep{corso24flexible} by defining our coupling distribution $q(\bm{z})$ as
\begin{equation}
    \label{equation:unbalanced_coupling_distribution}
    q(\bm{x}_{0}, \bm{x}_{1}) \propto q_{0}(\bm{x}_{0}) q_{1}(\bm{x}_{1}) \mathbb{I}_{c(\bm{x}_{0}, \bm{x}_{1}) \in c_{\mathbb{A}}},
\end{equation}
where $c_{\mathbb{A}}$ is defined as a set of apo-to-holo assessment filters measuring the structural similarity of the unbound (apo) and bound (holo) protein structures (n.b., not simply their binding pockets) in terms of their root mean square deviation (RMSD) and TM-score \citep{zhang2004scoring} following optimal structural alignment (as used in constructing PDBBind-E). Effectively, we sample independent examples from $q_{0}$ and $q_{1}$ and reject these paired examples if $c(\bm{x}_{0}, \bm{x}_{1}) < c_{\mathbb{A}_{TM}}$ or $c(\bm{x}_{0}, \bm{x}_{1}) \ge c_{\mathbb{A}_{RMSD}}$ (n.b., we use $c_{\mathbb{A}_{TM}} = 0.7$ and $c_{\mathbb{A}_{RMSD}} = 5$Å as well as other length-based criteria in our experiments, please see our code for full details).

\subsection{Sampling}
\label{section:sampling}
By default, we apply $i=40$ timesteps of an Euler solver to integrate \textsc{FlowDock}'s learned ODE $v_{\theta}$ forward in time for binding (holo) structure generation. Specifically, to generate structures, we propose to integrate a Variance Diminishing ODE (VD-ODE) that uses $v_{\theta}$ as
\begin{equation}
    \label{equation:sampling}
    \bm{x}_{n + 1} = clamp(\frac{1 - s}{1 - t} \cdot \eta) \cdot \bm{x}_{n} + clamp((1 - \frac{1 - s}{1 - t}) \cdot \eta) \cdot v_{\theta}(\bm{x}_{n}, t),
\end{equation}
where $n$ represents the current integer timestep, allowing us to define $t = \frac{n}{i}$ and $s = \frac{n + 1}{i}$; $\eta = 1.0$ in our experiments; and $clamp$ ensures both the LHS and RHS of Eq. \ref{equation:sampling} are lower and upper bounded by $1e^{-6}$ and $1 - 1e^{-6}$, respectively. We experimented with different values of $\eta$ yet ultimately settled on $1.0$ since this yielded \textsc{FlowDock}'s best performance for structure and affinity prediction. Intuitively, this VD-ODE solver limits the high levels of variance present in the model's predictions $v_{\theta}$ during early timesteps by sharply interpolating towards $v_{\theta}$ in later timesteps.

\section{Results}
\label{section:results}

\subsection{PoseBench protein-ligand docking}
\label{section:posebench_protein_ligand_docking}

\begin{figure*}[t]
    \centering
    \includegraphics[width=\textwidth, alt={Protein-ligand docking success rates of each baseline method on the PoseBusters Benchmark set (n=308). Error bars: 3 runs.}]{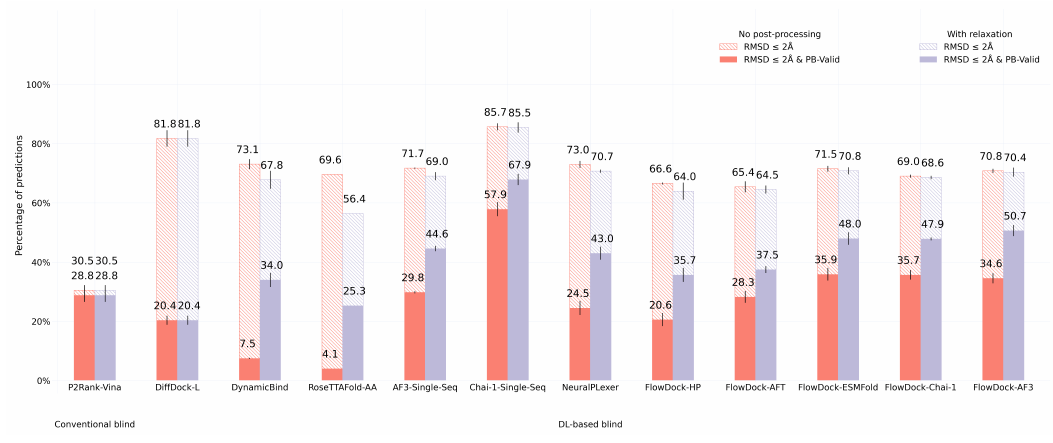}
    \caption{Protein-ligand docking success rates of each baseline method on the PoseBusters Benchmark set (n=308). Error bars: 3 runs.}
    \label{figure:posebusters_benchmark_bar_chart}
\end{figure*}

\begin{figure*}[t]
    \centering
    \includegraphics[width=0.95\textwidth, alt={Comparison of each flexible docking method's protein conformational changes made for the PoseBusters Benchmark set.}]{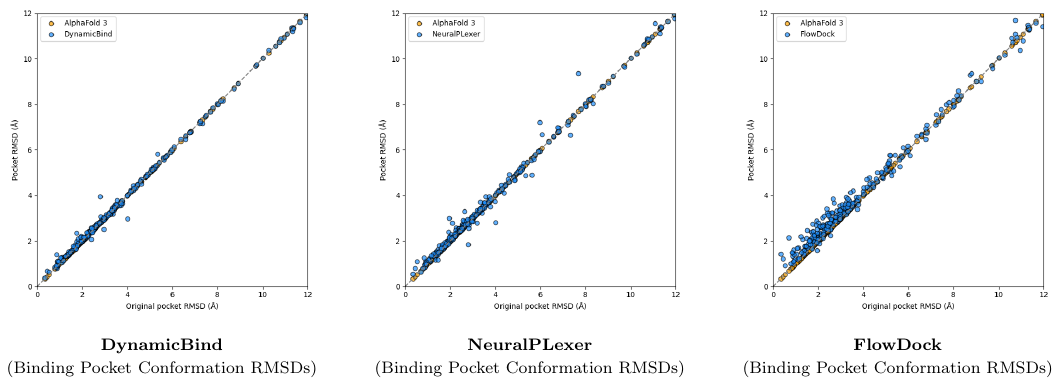}
    \caption{Comparison of each flexible docking method's protein conformational changes made for the PoseBusters Benchmark set (n=308).}
    \label{figure:posebusters_benchmark_protein_conformation_changes_scatter_plot}
\end{figure*}

\textbf{PoseBusters Benchmark set.} In Figures \ref{figure:posebusters_benchmark_bar_chart} and \ref{figure:posebusters_benchmark_protein_conformation_changes_scatter_plot}, we illustrate the performance of each baseline method for protein-ligand docking and protein conformational modification with the commonly-used PoseBusters Benchmark set \citep{buttenschoen2024posebusters}, provided by version 0.6.0 of the PoseBench protein-ligand benchmarking suite \citep{morehead2024deep}, which consists of 308 distinct protein-ligand complexes released after 2020. It is important to note that this benchmarking set can be considered a moderately difficult challenge for methods trained on recent collections of data derived from the Protein Data Bank (PDB) \citep{bank1971protein} such as PDBBind 2020 \citep{liu2015pdb}, as all of these 308 protein-ligand complexes are not contained in the most common training splits of such PDB-based data collections \citep{buttenschoen2024posebusters} (with the exception of AF3 which uses a cutoff date of September 30, 2021). Moreover, as described by \cite{buttenschoen2024posebusters}, a subset of these complexes also have very low protein sequence similarity to such training splits.

Figure \ref{figure:posebusters_benchmark_bar_chart} shows that \textsc{FlowDock} consistently improves over the original NeuralPLexer model's docking success rate in terms of its structural and chemical accuracy (as measured by the RMSD $\le$ 2Å \& PB-Valid metric \citep{buttenschoen2024posebusters}) and inter-run stability (as measured by the error bars listed). Notably, \textsc{FlowDock} achieves a 10\% higher docking success rate than NeuralPLexer without any structural energy minimization driven by molecular dynamics software \citep{eastman2017openmm}, and with energy minimization its docking success rate increases to 51\%, outperforming single-sequence AF3 and achieving second-best performance on this dataset compared to single-sequence Chai-1 \citep{chai2024chai}. Important to note is that Chai-1, like AF3, is a 10x larger model trained for one month using 128 NVIDIA A100 80GB GPUs on more than twice as much data in the PDB deposited up to 2021, whereas \textsc{FlowDock} is trained using only 4 80GB H100 GPUs for one week, representing a 32x reduction in GPU hours required for training. Additionally, \textsc{FlowDock} outperforms the \textit{hybrid} flexible docking method DynamicBind \citep{lu2024dynamicbind} by more than 16\%, which is a comparable model in terms of its size, training, and downstream capabilities for drug discovery. Our results with ablated versions of \textsc{FlowDock} trained instead with a protein harmonic prior (\textsc{FlowDock-HP}) or with affinity prediction frozen until a fine-tuning phase (\textsc{FlowDock-AFT}) highlight that the protein ESMFold prior the base \textsc{FlowDock} model employs has imbued it with meaningful structural representations for accurate ligand binding structure prediction that are robust to changes in the source method of \textsc{FlowDock}'s predicted protein input structures (e.g., \textsc{FlowDock-ESMFold} vs. \textsc{FlowDock-Chai-1} vs. \textsc{FlowDock-AF3}), providing users with multiple structure prediction options (e.g., ESMFold for faster and commercially available prediction inputs).

A surprising finding illustrated in Figure \ref{figure:posebusters_benchmark_protein_conformation_changes_scatter_plot} is that no method can consistently improve the binding pocket RMSD of AF3's initial protein structural conformations, which contrasts with the results originally reported for flexible docking methods such as DynamicBind which used structures predicted by AF2 \citep{jumper2021highly} in its experiments. From this figure, we observe that DynamicBind and NeuralPLexer both infrequently modify AF3's initial binding pocket structure, whereas \textsc{FlowDock} often modifies the pocket structure during ligand binding. The former two methods occasionally improve largely-correct initial pocket conformations by $\sim$1Å, whereas \textsc{FlowDock} primarily does so for mostly-incorrect initial pockets.

\begin{figure*}[t]
    \centering
    \includegraphics[width=\textwidth, alt={Protein-ligand docking success rates of each baseline method on the DockGen-E set (n=14). Error bars: 3 runs.}]{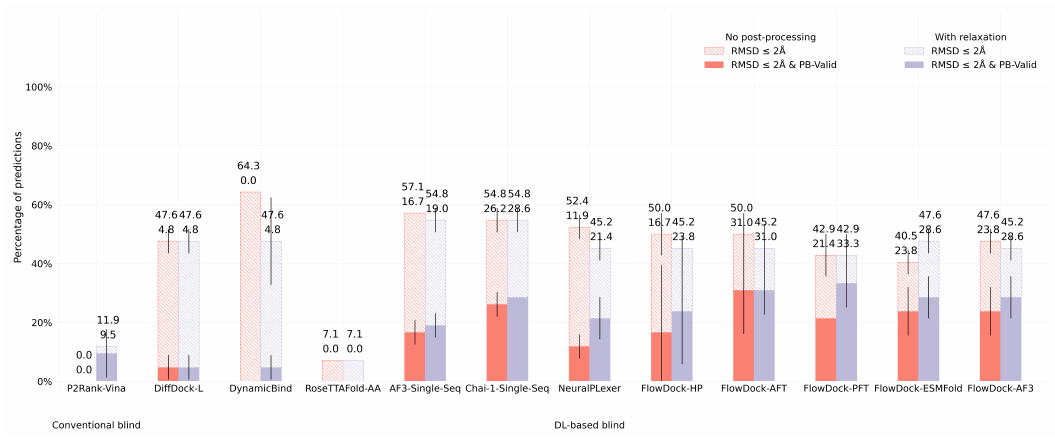}
    \caption{Protein-ligand docking success rates of each baseline method on the DockGen-E set (n=14). Error bars: 3 runs.}
    \label{figure:dockgen_bar_chart}
\end{figure*}

\begin{figure*}[t]
    \centering
    \includegraphics[width=0.95\textwidth, alt={Comparison of each flexible docking method's protein conformational changes made for the DockGen-E set.}]{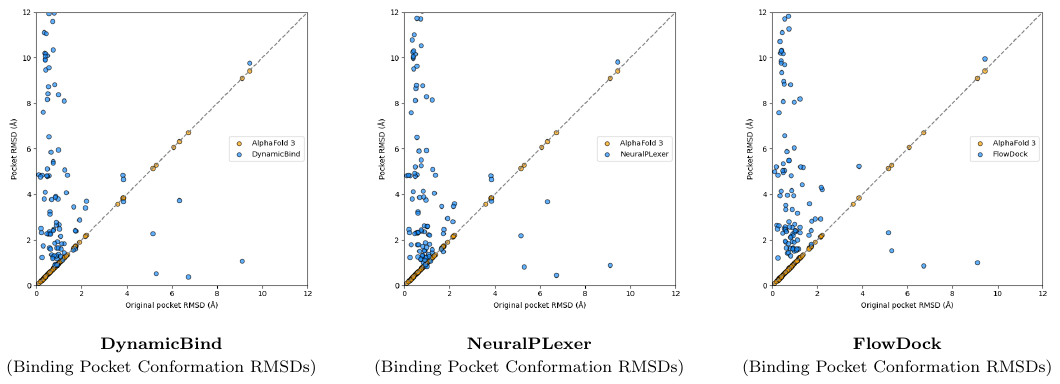}
    \caption{Comparison of each flexible docking method's protein conformational changes made for the DockGen-E set (n=122).}
    \label{figure:dockgen_protein_conformation_changes_scatter_plot}
\end{figure*}

\textbf{DockGen-E set.} To assess the generalization capabilities of each baseline method, in Figures \ref{figure:dockgen_bar_chart} and \ref{figure:dockgen_protein_conformation_changes_scatter_plot}, we report each method's protein-ligand docking and protein conformational modification performance for the novel (i.e., naturally rare) protein binding pockets found in the new DockGen-E dataset from PoseBench. Each of DockGen-E's protein-ligand complexes represents a distinct binding pocket that facilitates a unique biological function described by its associated ECOD domain identifier \citep{corsodeep}. As our results for the DockGen-E dataset show in Figure \ref{figure:dockgen_bar_chart}, most DL-based docking or structure prediction methods have likely not been trained or overfitted to these binding pockets, as this dataset's best docking success rate achieved by any method is approximately 33\%, much lower than the 68\% best docking success rate achieved for the PoseBusters Benchmark set. We find further support for this phenomenon in Figure \ref{figure:dockgen_protein_conformation_changes_scatter_plot}, where we see that all DL-based flexible docking methods find it challenging to avoid degrading the initial binding pocket state predicted by AF3 yet all methods can \textit{restore} a handful of AF3 binding pockets to their bound (holo) form. This suggests that all DL methods (some more so than others) struggle to generalize to novel binding pockets, yet \textsc{FlowDock} achieves top performance in this regard by tying with single-sequence Chai-1. Further, to address this generalization issue, our preliminary results fine-tuning \textsc{FlowDock} for 48 hours using the new, diverse PLINDER \citep{durairaj24plinder} dataset (i.e., \textsc{FlowDock-PFT}), where we use the dataset's crystal apo-to-holo mapped protein-ligand complex structures contained within its default PL50 training split and deposited in the PDB before 2018, suggest that comprehensively training new DL methods on diverse protein-ligand binding structures is a promising direction towards generalizable docking.

\begin{table}
  \caption{\textbf{Computational resource requirements.} The average structure prediction runtime (in seconds) and peak memory usage (in GB) of baseline methods on a 25\% subset of the Astex Diverse dataset \citep{hartshorn2007diverse} using an NVIDIA 80GB A100 GPU for benchmarking (with baselines taken from \citep{morehead2024deep}). The symbol \textsc{-} denotes a result that could not be estimated.}
  \label{table:baseline_method_average_compute_resources}
  \centering
  \resizebox{\columnwidth}{!}{%
      \begin{tabular}{lccc}
        \toprule
        Method    & Runtime (s)     &   CPU Memory Usage (GB)   &   GPU Memory Usage (GB)    \\
        \midrule
        P2Rank-Vina  &  1,283.70   &   9.62    &   0.00    \\
        DiffDock-L  &   88.33  &   8.99    &   70.42   \\
        DynamicBind & 146.99 &   5.26    &   18.91   \\
        RoseTTAFold-All-Atom  & 3,443.63  &   55.75   &   72.79   \\
        AF3 & 3,049.41  &   -   &   -   \\
        AF3-Single-Seq & 58.72  &   -   &   -   \\
        Chai-1-Single-Seq & 114.86 &   58.49   &   56.21   \\
        NeuralPLexer  &    29.10  &   11.19   &   31.00   \\
        \rowcolor[gray]{0.8} \textbf{\textsc{FlowDock}}  &    39.34  &   11.98   &   25.61   \\
        \bottomrule
      \end{tabular}
    }
\end{table}

\textbf{Computational resources.} To formally measure the computational resources required to run each baseline method, in Table \ref{table:baseline_method_average_compute_resources} we list the average runtime (in seconds) and peak CPU (GPU) memory usage (in GB) consumed by each method when running them on a 25\% subset of the Astex Diverse dataset \citep{hartshorn2007diverse} (baseline results taken from \cite{morehead2024deep}). Here, we notably find that \textsc{FlowDock} provides the second lowest computational runtime and GPU memory usage compared to all other DL methods, enabling one to use commodity computing hardware to quickly screen new drug candidates using combinations of \textsc{FlowDock}'s predicted heavy-atom structures, confidence scores, and binding affinities.

\begin{figure*}[t]
    \centering
    \includegraphics[width=\textwidth, alt={Comparison of \textsc{DynamicBind} and \textsc{FlowDock}'s predicted structures (w/o hydrogens) and crystal PDBBind test example 6I67.}]{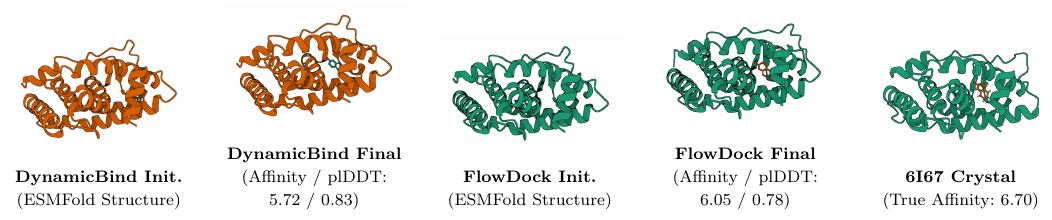}
    \caption{Comparison of \textsc{DynamicBind} and \textsc{FlowDock}'s predicted structures (w/o hydrogens) and crystal PDBBind test example 6I67.}
    \label{figure:pdbbind_6i67_example}
\end{figure*}

\subsection{PDBBind binding affinity estimation}
\label{section:pdbbind_binding_affinity_estimation}

\begin{table}
    \caption{\textbf{Binding affinity estimation using PDBBind test set.} For all methods, binding affinities were predicted in \textit{one shot} using the commonly-used 363 PDBBind (ligand and time-split) test complexes (with splits and baselines from \citet{lu2024dynamicbind}). Results for \textsc{FlowDock} are reported as the mean and standard error of measurement ($n = 3$) of each metric over three independent runs.
    Note that, for historical reasons, the results for each version of \textsc{FlowDock} were obtained using ESMFold predicted protein input structures.
    }
    \label{table:pdbbind_binding_affinity_estimation}
    \centering
    \resizebox{\columnwidth}{!}{%
        \begin{tabular}{c|c|c|c|c}
        \hline
        \multicolumn{1}{c}{Method} & \multicolumn{1}{c}{Pearson ($\uparrow$)} & \multicolumn{1}{c}{Spearman ($\uparrow$)} & \multicolumn{1}{c}{RMSE ($\downarrow$)} & \multicolumn{1}{c}{MAE ($\downarrow$)}  \\ \hline
        GIGN & 0.286 & 0.318 & 1.736 & 1.330 \\
        TransformerCPI & 0.470 & 0.480 & 1.643 & 1.317 \\
        MONN & 0.545 & 0.535 & 1.371 & 1.103 \\
        TankBind & 0.597 & 0.610 & 1.436 & 1.119 \\
        DynamicBind (One-Shot) & 0.665 & 0.634 & 1.301 & 1.060 \\ \hline
        \rowcolor[gray]{0.8} \textsc{FlowDock-HP} & $0.577 \pm 0.001$ & $0.560 \pm 0.001$ & $1.516 \pm 0.001$ & $1.196 \pm 0.002$ \\
        \rowcolor[gray]{0.8} \textsc{FlowDock-AFT} & $0.663 \pm 0.003$ & $0.624 \pm 0.003$ & $1.392 \pm 0.005$ & $1.113 \pm 0.003$ \\
        \rowcolor[gray]{0.8} \textbf{\textsc{FlowDock}} & $\mathbf{0.705 \pm 0.001}$ & $\mathbf{0.674 \pm 0.002}$ & \underline{1.363$\pm 0.003$} & \underline{1.067$\pm 0.003$} \\ \hline
        \end{tabular}
    }
\end{table}

In this section, we explore binding affinity estimation with \textsc{FlowDock} using the PDBBind 2020 test dataset (n=363) originally curated by \citep{stark2022equibind}, with benchmarking results shown in Table \ref{table:pdbbind_binding_affinity_estimation}. Popular affinity prediction baselines listed in Table \ref{table:pdbbind_binding_affinity_estimation} such as TankBind \citep{lu2022tankbind} and DynamicBind \citep{lu2024dynamicbind} demonstrate that accurate affinity estimations are possible using hybrid DL models of protein-ligand structures and affinities. Here, we find that, as a hybrid deep generative model, \textsc{FlowDock} provides the best Pearson and Spearman's correlations compared to all other baselines including
\textsc{FlowDock-HP} (a fully harmonic variant of \textsc{FlowDock}) and \textsc{FlowDock-AFT} (an ESMFold prior variant trained first for structure prediction and then with affinity fine-tuning)
and produces compelling root mean squared error (RMSE) and mean absolute error (MAE) rates compared to the previous state-of-the-art method DynamicBind. Referencing Table \ref{table:baseline_method_average_compute_resources}, we further note that \textsc{FlowDock}'s average computational runtime per protein-ligand complex is more than 3 times lower than that of DynamicBind, demonstrating that \textsc{FlowDock}, to our best knowledge, is currently the \textit{fastest} binding affinity estimation method to match or exceed DynamicBind's level of accuracy for predicting binding affinities using the PDBBind 2020 dataset.

In Figure \ref{figure:pdbbind_6i67_example}, we provide an illustrative example of a protein-ligand complex in the PDBBind test set (6I67) for which \textsc{FlowDock} predicts notably more accurate complex structural motions and binding affinity values than the hybrid DL method DynamicBind, importantly recognizing that the right-most protein loop domain should be moved further to the right to facilitate ligand binding (see Appendix \ref{appendix:structure_generation_example_trajectory} of our Supplementary Materials for an example of one of \textsc{FlowDock}'s interpretable structure generation trajectories). One should note that, for historical reasons, our experiments with this PDBBind-based test set employed protein structures predicted by ESMFold (not AF3). In the next section, we explore an even more practical application of \textsc{FlowDock}'s fast and accurate structure and binding affinity predictions in the CASP16 ligand prediction competition.

\subsection{CASP16 protein-ligand binding affinity prediction}
\label{section:casp16_protein_ligand_binding_affinity_prediction}

\begin{figure*}[t]
    \centering
    \includegraphics[width=\textwidth, alt={Protein-ligand binding affinity prediction rankings for the CASP16 ligand prediction category (n=140).}]{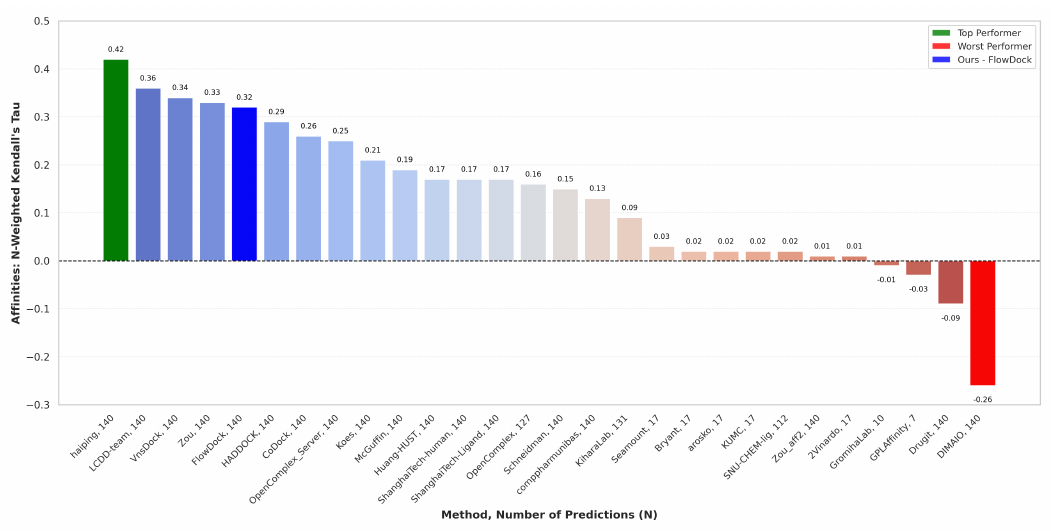}
    \caption{Protein-ligand binding affinity prediction rankings for the CASP16 ligand prediction category (n=140).}
    \label{figure:casp16_binding_affinity_rankings_plot}
\end{figure*}

In Figure \ref{figure:casp16_binding_affinity_rankings_plot}, we illustrate the performance of each predictor group for blind protein-ligand binding affinity prediction in the ligand category of the CASP16 competition held in summer 2024, in which pharmaceutically relevant binding ligands were the primary focus of this competition. Notably, \textsc{FlowDock} is the \textit{only} hybrid (structure \& affinity prediction) ML method represented among the top-5 predictors, demonstrating the robustness of its knowledge of protein-ligand interactions. Namely, all other top prediction methods were trained specifically for binding affinity estimation assuming a predicted or crystal complex structure is provided. In contrast, in CASP16, we demonstrated the potential of using \textsc{FlowDock} to predict \textit{both} protein-ligand structures and binding affinities and using its top-5 predicted structures' structural confidence scores to rank-order its top-5 binding affinity predictions (see Appendices \ref{appendix:casp16_structure_prediction_results} and \ref{appendix:posebusters_benchmark_ligand_dissimilarity_structure_prediction_results} of our Supplementary Materials for \textsc{FlowDock}'s e.g., CASP16 structure prediction results). Ranked 5th for binding affinity estimation, these results of the CASP16 competition demonstrate that this dual approach of predicting protein-ligand structures and binding affinities with a single DL model (\textsc{FlowDock}) yields compelling performance for virtual screening of pharmaceutically interesting molecular compounds.

\section{Conclusion}

In this work, we have presented \textsc{FlowDock}, a novel, state-of-the-art deep generative flow model for fast and accurate (hybrid) protein-ligand binding structure and affinity prediction. Benchmarking results suggest that \textsc{FlowDock} achieves structure prediction results better than single-sequence AF3 and comparable to single-sequence Chai-1 and outperforms existing hybrid models like DynamicBind across a range of binding ligands. Lastly, we have demonstrated the pharmaceutical virtual screening potential of \textsc{FlowDock} in the CASP16 ligand prediction competition, where it achieved top-5 performance. Future work could include retraining the model on larger and more diverse clusters of protein-ligand complexes, experimenting with new ODE solvers, or scaling up its parameter count to see if it displays any scaling law behavior for structure or affinity prediction. As a deep generative model for structural biology made available under an MIT license, we believe \textsc{FlowDock} takes a notable step forward towards fast, accurate, and broadly applicable modeling of protein-ligand interactions.


\section{Conflict of interest}
No conflicts of interest are declared.

\section{Author contributions statement}

A.M. and J.C. conceived the research. A.M. conducted the experiment(s). J.C. acquired funding to support this work. A.M. and J.C. analyzed the results and wrote the manuscript.

\section{Funding}
The authors thank the anonymous reviewers for their valuable suggestions. This work was supported by a U.S. NSF grant (DBI2308699) and a U.S. NIH grant (R01GM093123) awarded to J.C. Additionally, this work was performed using computing infrastructure provided by Research Support Services at the University of Missouri-Columbia (DOI: 10.32469/10355/97710).


\begin{appendices}

\section{Geometric flow matching training and inference}
\label{appendix:pseudocode}

We characterize \textsc{FlowDock}'s training and sampling procedures in Sections \ref{section:training} (Training) and \ref{section:sampling} (Sampling) of the main text, respectively. To further illustrate how training and inference with \textsc{FlowDock} work, in Algorithms \ref{algorithm:training} and \ref{algorithm:inference} we provide the corresponding pseudocode. For more details, please see our accompanying source code at \url{https://github.com/BioinfoMachineLearning/FlowDock}.

\begin{algorithm}
\caption{Training}
\label{algorithm:training}
\begin{algorithmic}[1]
\Require Training examples of binding site-aligned apo (holo) protein (ligand) structures, protein sequences, ligand SMILES strings, and binding affinities $\{(X_{a_{i}}^{P}, X_{h_{i}}^{P}, X_{h_{i}}^{L}, S_i, M_i, B_i)\}$
\ForAll{$(X_{a_{i}}^{P}, X_{h_{i}}^{P}, X_{h_{i}}^{L}, S_i, M_i, B_i)$}
    \State Extract $x_{1}^{P}, x_{1}^{L} \gets \text{HeavyAtoms}(X_{h_{i}}^{P}, X_{h_{i}}^{L})$
    \State Sample $x_0^{P} \gets \text{ESMFold}(S_i) + \epsilon, \hspace{0.25cm} \epsilon \sim \mathcal{N}(0, \sigma=1e^{-4})$
    \State Sample $x_0^{L} \gets \text{HarmonicPrior}(M_{i_{frag}}), \hspace{0.25cm} \forall frag \in M_i$
    \State Sample $t \sim \mathcal{U}(0, 1)$
    \State Concatenate $x_{0} = \text{Concat}(x_{0}^{P}, x_{0}^{L})$
    \State Concatenate $x_{1} = \text{Concat}(x_{1}^{P}, x_{1}^{L})$
    \State Interpolate $x_t \gets t \cdot x_1 + (1 - t) \cdot x_0$
    \State Predict $\hat{X}_{h_{i}} \gets \text{NeuralPLexer}(S_i, M_i, x_t, t)$
    \State Predict $\hat{B}_i \gets \text{ESDM}_{aff}(S_i, M_i, \text{StopGrad}(\hat{X}_{h_{i}}))$
    \State Optimize losses $\mathcal{L}_{X} := \lambda_{X} \cdot \text{FAPE}(X_{h_{i}}, \hat{X}_{h_{i}}) + \mathcal{L}_{B} := \lambda_{B} \cdot \text{MSE}(\hat{B}_i, B_i), \hspace{0.25cm} \lambda_{X} = 0.2, \hspace{0.25cm} \lambda_{B} = 0.1$
\EndFor
\end{algorithmic}
\end{algorithm}

\begin{algorithm}
\caption{Inference}
\label{algorithm:inference}
\begin{algorithmic}[1]
\Require Protein sequences and ligand SMILES strings $(S, M)$
\Ensure Sampled top-5 heavy-atom structures $\hat{X}$ with confidence scores $\hat{C}$ and binding affinities $\hat{B}$
\State Sample $x_0^{P} \gets \text{ESMFold}(S) + \epsilon, \hspace{0.25cm} \epsilon \sim \mathcal{N}(0, \sigma=1e^{-4})$
\State Sample $x_0^{L} \gets \text{HarmonicPrior}(M_{frag}), \hspace{0.25cm} \forall frag \in M$
\State Concatenate $x_{0} = \text{Concat}(x_{0}^{P}, x_{0}^{L})$
\For{$n \gets 0 \text{ to } i$}
    \State Let $t \gets \frac{n}{i}$ and $s \gets \frac{n + 1}{i}$
    \State Predict $\hat{X} \gets \text{NeuralPLexer}(S, M, x_n, t)$
    \If{$n = i - 1$}
        \State Predict $\hat{C} \gets \text{ESDM}_{conf}(S, M, \hat{X})$ \hspace{0.25cm} \# Pre-trained
        \State Predict $\hat{B} \gets \text{ESDM}_{aff}(S, M, \hat{X})$
        \State Rank top-5 $\hat{X}$ and $\hat{B}$ using $\hat{C}$
        \State \Return $\hat{X}, \hat{C}, \hat{B}$
    \EndIf
    \State Extract $\hat{x}_1 \gets \text{HeavyAtoms}(\hat{X})$
    \State Align $x_n \gets \text{RMSDAlign}(x_n, \hat{x}_1)$
    \State Interpolate $x_{n + 1} = clamp(\frac{1 - s}{1 - t} \cdot \eta) \cdot x_{n} + clamp((1 - \frac{1 - s}{1 - t}) \cdot \eta) \cdot \hat{x}_1, \hspace{0.25cm} \eta = 1$
\EndFor
\end{algorithmic}
\end{algorithm}

\clearpage

\section{Structure generation example trajectory}
\label{appendix:structure_generation_example_trajectory}

\begin{figure*}[t]
    \centering
    \includegraphics[width=\textwidth, alt={Comparison of \textsc{FlowDock}'s predicted structure states (w/o hydrogens) for CASP16 superligand pose pharma target L3008.}]{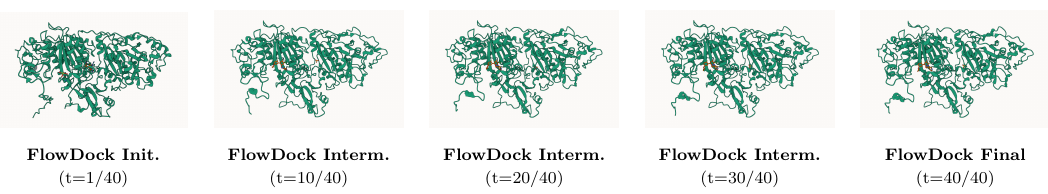}
    \caption{Comparison of \textsc{FlowDock}'s predicted structure states (w/o hydrogens) for CASP16 superligand pose pharma target L3008.}
    \label{figure:casp16_l3008_example}
\end{figure*}

To illustrate one of \textsc{FlowDock}'s interpretable structure generation trajectories using conditional flow matching, in Figure \ref{figure:casp16_l3008_example}, we report \textsc{FlowDock}'s predicted structural states for CASP16 superligand pose pharma target L3008, notably a \textit{multi}-ligand pose target, in evenly spaced increments throughout \textsc{FlowDock}'s generation trajectory. In short, we see that \textsc{FlowDock} enables multi-ligand protein complexes to be predicted through concise flow trajectories, yielding early protein and ligand conformational changes following the model's initial binding pocket prediction.

\section{CASP16 structure prediction results}
\label{appendix:casp16_structure_prediction_results}

\begin{figure*}[t]
    \centering
    \includegraphics[width=\textwidth, alt={Comparison of the protein-ligand structure prediction results of \textsc{FlowDock} and the deep learning ensembling method MULTICOM\_ligand in terms of their binding pocket-aligned ligand RMSDs for the CASP16 superligand pose pharma targets (n=301).}]{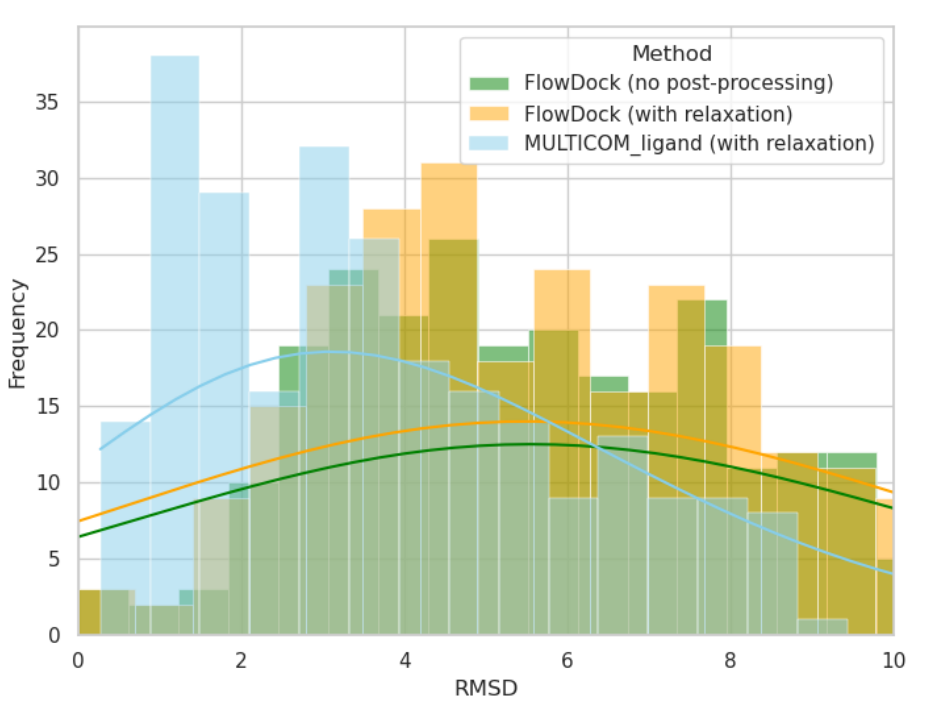}
    \caption{Comparison of the protein-ligand structure prediction results of \textsc{FlowDock} and the deep learning ensembling method MULTICOM\_ligand in terms of their binding pocket-aligned ligand RMSDs for the CASP16 superligand pose pharma targets (n=301).}
    \label{figure:casp16_structure_prediction_results}
\end{figure*}

\begin{figure*}[t]
    \centering
    \includegraphics[width=\textwidth, alt={Comparison of the protein-(multi-)ligand structure prediction results of \textsc{FlowDock} and the deep learning ensembling method MULTICOM\_ligand in terms of their binding pocket-aligned ligand RMSDs for the CASP16 superligand pose pharma targets (n=126).}]{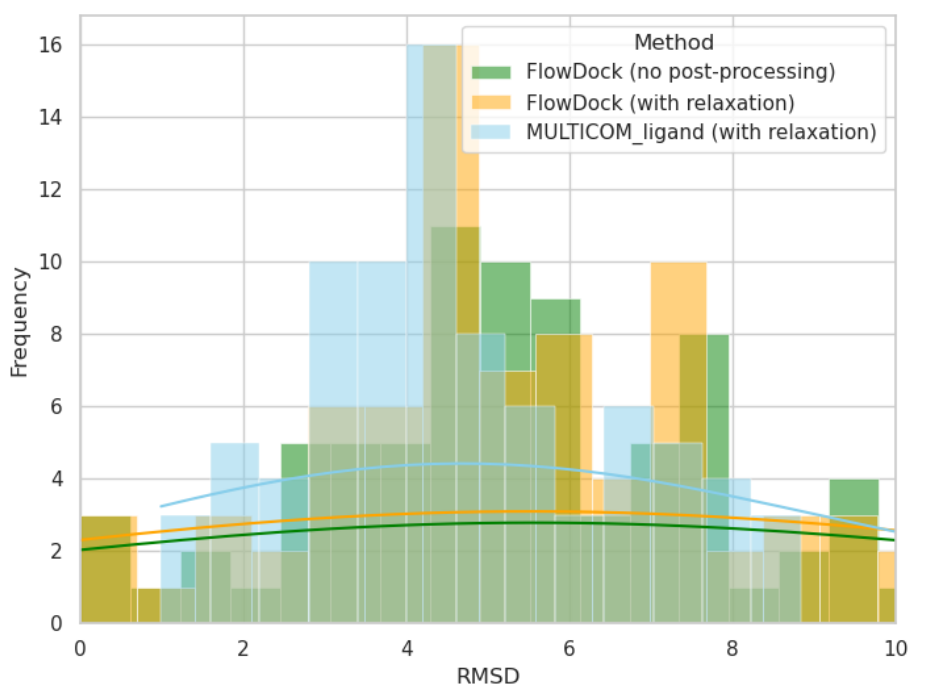}
    \caption{Comparison of the protein-(multi-)ligand structure prediction results of \textsc{FlowDock} and the deep learning ensembling method MULTICOM\_ligand in terms of their binding pocket-aligned ligand RMSDs for the CASP16 superligand pose pharma targets (n=126).}
    \label{figure:casp16_multi_ligand_structure_prediction_results}
\end{figure*}

In Figure \ref{figure:casp16_structure_prediction_results}, we compare the protein-ligand structure prediction RMSDs of \textsc{FlowDock} and MULTICOM\_ligand \citep{morehead2025protein}, a top-5 multi-model deep learning prediction method in the CASP16 ligand prediction category, for the 231 superligand pose pharma targets made available during the 16th Critical Assessment of Techniques for Structure Prediction (CASP16). As these results demonstrate, \textsc{FlowDock}, as a standalone deep learning method, achieves competitive structure predictions for many of the new CASP16 ligand targets. Similarly, Figure \ref{figure:casp16_multi_ligand_structure_prediction_results} illustrates that \textsc{FlowDock} and MULTICOM\_ligand are approximately tied in terms of their ability to structurally model CASP16's 56 \textit{multi}-ligand protein complexes, further highlighting the broad applicability of \textsc{FlowDock}'s structure predictions in diverse drug discovery settings.

\clearpage

\clearpage

\section{PoseBusters Benchmark ligand dissimilarity structure prediction results}
\label{appendix:posebusters_benchmark_ligand_dissimilarity_structure_prediction_results}

\begin{figure*}[t]
    \centering
    \includegraphics[width=\textwidth, alt={Analysis of the protein-ligand structure prediction results of \textsc{FlowDock} in terms of its binding pocket-aligned ligand RMSDs for the chemically dissimilar (multi-)ligand PoseBusters Benchmark targets (n=18).}]{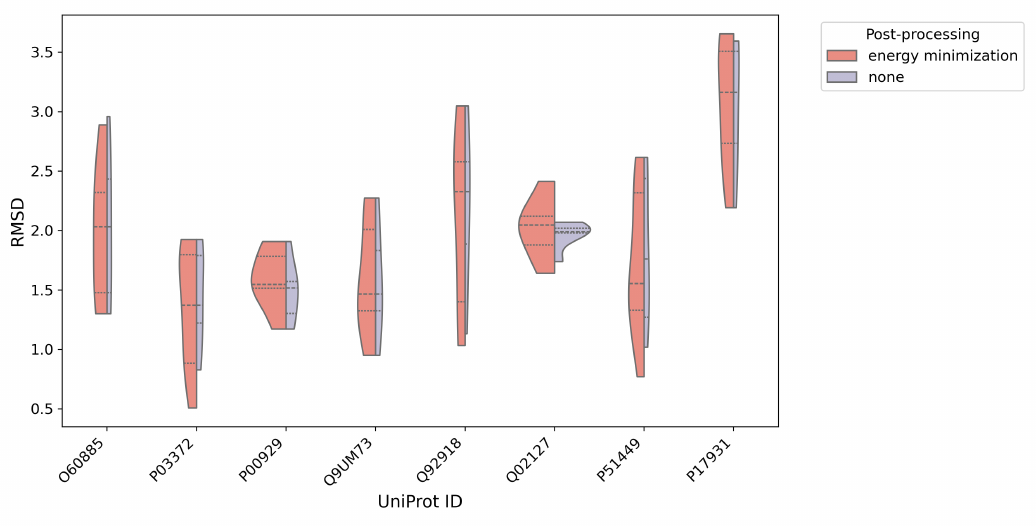}
    \caption{Analysis of the protein-ligand structure prediction results of \textsc{FlowDock} in terms of its binding pocket-aligned ligand RMSDs for the chemically dissimilar (multi-)ligand PoseBusters Benchmark targets (n=18).}
    \label{figure:posebusters_benchmark_ligand_dissimilarity_structure_prediction_results}
\end{figure*}

To investigate \textsc{FlowDock}'s chemical generalization capabilities, in Figure \ref{figure:posebusters_benchmark_ligand_dissimilarity_structure_prediction_results}, we illustrate the structure prediction performance of \textsc{FlowDock} for chemically dissimilar (Tanimoto similarity $<$ 0.6) ligands associated with the same protein target in the PoseBusters Benchmark dataset. Figure \ref{figure:posebusters_benchmark_ligand_dissimilarity_structure_prediction_results} shows that \textsc{FlowDock}'s average ligand RMSD of each of these (multi-)ligand protein targets is approximately 2 \AA, with a standard deviation around 1 \AA, highlighting that its predictions for chemically dissimilar intra-protein ligands are of high average accuracy and demonstrate generalizability with the consistency of \textsc{FlowDock}'s average inter-ligand RMSD differences.

\end{appendices}

\end{document}